\documentclass{bmvc2k}

\title{ExpertNet: Adversarial Learning and Recovery Against Noisy Labels}

\addauthor{Amirmasoud Ghiassi}{s.ghiassi@tudelft.nl}{1}
\addauthor{Robert Birke}{robert.birke@ch.abb.com}{2}
\addauthor{Rui Han}{hanrui@bit.edu.cn}{3}
\addauthor{Lydia Y.Chen}{lydiaychen@ieee.org}{1}

\addinstitution{
 TU Delft \\
 The Netherlands
}
\addinstitution{
 ABB Future Labs\\
 Switzerland
}
\addinstitution{
Beijing Institute of Technology\\
China
}
\runninghead{Ghiassi, Birke, Chen}{ExpertNet}

\def\ENet{Expert\xspace}
\def\ANet{Amateur\xspace}
\def\system{ExpertNet\xspace}

\usepackage{graphicx}
\usepackage{amsmath,amssymb} 
\usepackage{soul}
\usepackage{amsmath}
\usepackage{amssymb}
\usepackage[T1]{fontenc}   
\usepackage{xspace}
\usepackage[vlined,ruled,linesnumbered]{algorithm2e}
\usepackage{url}
\usepackage{bm} 
\usepackage{makecell}
\usepackage{diagbox}
\usepackage{multirow}
\usepackage{multicol}
\usepackage{lipsum}                     
\usepackage{xargs}                      

\usepackage{enumitem}
\begin{document}

\maketitle

\begin{abstract}
   Today's available datasets in the wild, e.g., from social media and open platforms, present tremendous opportunities and challenges for deep learning, as there is a significant portion of tagged images, but often with noisy, i.e. erroneous, labels. Recent studies improve the robustness of deep models against noisy labels without the knowledge of true labels. In this paper, we advocate to derive a stronger classifier which proactively makes use of the noisy labels in addition to the original images - turning noisy labels into learning features. To such an end, we propose a novel framework, \system, composed of \ANet and \ENet, which iteratively learn from each other. \ANet is a regular image classifier trained by the feedback of \ENet, which imitates how human experts would correct the predicted labels from \ANet using the noise pattern learnt from the knowledge of both the noisy and ground truth labels. The trained \ANet and \ENet proactively leverage the images and their noisy labels to infer image classes. Our empirical evaluations on noisy versions of CIFAR-10, CIFAR-100 and real-world data of Clothing1M show that the proposed model can achieve robust classification against a wide range of noise ratios and with as little as 20-50\% training data, compared to state-of-the-art deep models that solely focus on distilling the impact of noisy labels.
\end{abstract}

\section{Introduction} 
The ever-increasing self-generated contents on social media, e.g., Instagram images, power up the deep neural networks, but also aggravate the challenge of noisy data. Large portion of images accessible on the public domain come with labels which are unfortunately noisy due to careless annotations~\cite{yan2014learning,blum2003noise} or even adversary strategies~\cite{Szegedy2014intriguing,kurakin2016adversarial,goodfellow2014explaining}. 
The learning capacity of deep neural networks is shown to be hindered by such noisy labels~\cite{Zhang2017memorization} due to the memorization effect of networks. Classification accuracy on standard image benchmarks degrades drastically in the presence of dirty labels. For example, the accuracy of a trained AlexNet to classify CIFAR-10 images drops from 77\% to 10\%, when trained on noisy labels~\cite{Zhang2017memorization}.


Motivated by the significant impact of noisy labels, the prior art~\cite{jiang2017mentornet} derives different robust deep networks with a central theme to distill the influence of noisy labels in the model training process without the knowledge of the label ground truth. 
As a result, the learned networks can robustly classify images in a stand-alone manner. D2L~\cite{ma2018dimensionality} estimates the Local Intrinsic Dimension (LID) at each epoch as a proxy to indicate the existence and impact of dirty labels. Co-teaching~\cite{han2018co} trains two networks simultaneously by exchanging the weights updated from possibly clean data.  Forward~\cite{patrini2017making} uses a noise-aware correction matrix to correct labels and train the network. Bootstrap~\cite{reed2014training} has a loss function which combines predicted and noisy labels.
 
While prior art significantly improves the robustness of deep networks, the pre-assumed scenarios overlook the opportunity of noisy labels.
On the one hand, today's image data are often bundled with labels of questionable quality and detrimental impact on the learning. On the other hand, such labels provide auxiliary information which can compliment the learnt knowledge of deep networks trained solely on image inputs. The core idea behind visual-semantic models, e.g., DeVise~\cite{Frome:NIPS13:DeVise}, is to combine the learning capacities of labeled images and annotated data.
C-GAN~\cite{mirza2014conditional} (conditional generative adversarial network) improves the quality of images synthesized by the generator network via additional label information and RC-GAN~\cite{thekumparampil2018robustness} further addresses the challenge of dirty labels for C-GAN.

In this paper, we advocate to leverage the noisy labels as an additional feature to derive a stronger classifier. 
We consider learning scenarios where at training time both the ground truth and noisy labels are available, and only noisy labels at inference time. In particular difficult classification problems, whose labels require a high degree of expertise, can fit this scenario well. One such example is cancer detection from medical images. This is a daunting task, and even trained experts are prone to make errors. Hence, these images are evaluated by multiple doctors of varying expertise. In such a setting, both noisy (first evaluation by one expert) and true labels (e.g., stemming from a committee or subsequent in-depth exams) are available at the same time.

We derive a robust network, namely \system, composed of \ANet and \ENet, where the former classifies images based on the feedback from \ENet and the latter learns how to correct the output of \ANet like human experts. Both models are trained simultaneously at each minibatch. \ANet learns to classify the input images to the corrected labels from \ENet, and the softmax output of \ANet plus the given labels are inputs to train \ENet to match the ground truth. \ANet can be seen as a regular image classifier, which \ENet helps it to be aware of the presence of noisy labels. Such trained \ANet and \ENet can then classify images based on the image and corresponding noisy label.

We empirically evaluate \system on synthetic noise injected into CIFAR-10 and CIFAR-100, and noise drawn from real world contained in Clothing1M. For a fair comparison with state-of-the-art robust deep models, we present the classification accuracy in both \ANet only and complete \system model under different subsets of training inputs. \system consistently outperforms existing image-only models, i.e., D2L, Co-teaching, Forward and Boostrap, especially for CIFAR-100. When using the same amount of training data, \system can achieve absolute accuracy improvements of 5\% up to 30\%. \system reaches similar or higher accuracy than image-only models even with just 20\% of training data in the case of CIFAR-100. 
 
Our contribution can be summarized as follows. First, we derive a novel network framework, i.e., \system, that turns noisy labels into auxiliary learning advantages via imitating human experts. Secondly, we significantly improve the robustness of deep networks against noisy labels compared to models based on images only.



\subsection{Problem statement}
\begin{figure*}[htp]
    \includegraphics[trim={0 0.2cm 0 0.35cm},clip,width=0.6\linewidth]{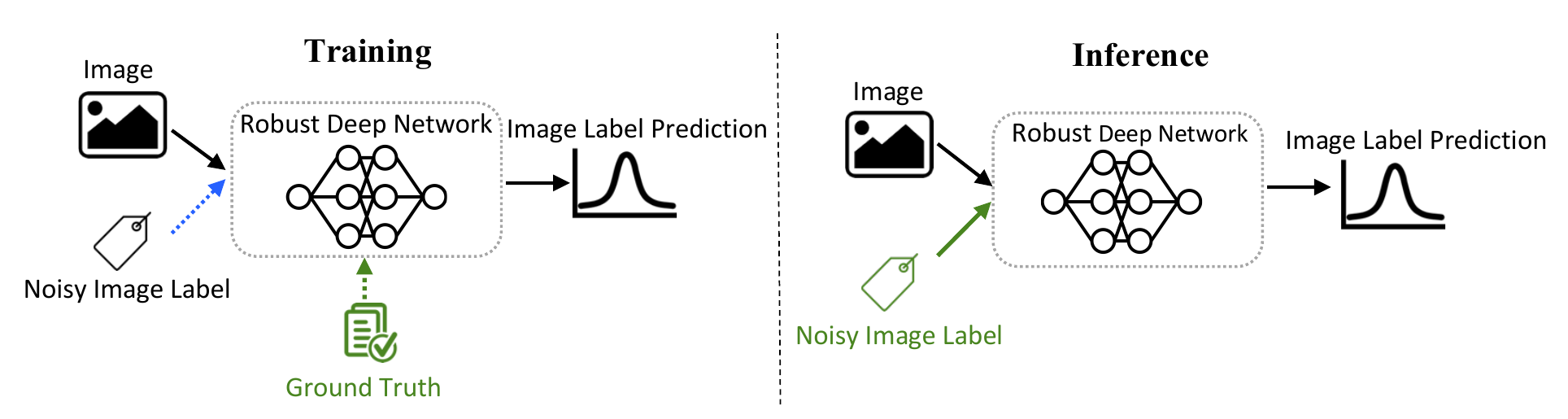}
	\centering
\caption{Training and testing image classifiers with noisy labels.}
\label{fig:TTmodels}
\end{figure*}


The problem considered here is as follows. Images collected in the public domain are tagged with pre-existing noisy labels, whose true classes can be corrupted. We assume label noises follow random distribution. 
We illustrate in Fig.~\ref{fig:TTmodels} (black elements only) the learning procedure that is commonly deployed by robust deep networks~\cite{ma2018dimensionality,han2018co,patrini2017making,reed2014training}. The deep networks are trained by a set of images and labels, which are noisy, meaning with incorrect label classes.
The objective of the training process is to minimize the loss function, which may be modified to be noise tolerant~\cite{ma2018dimensionality}. The network architecture may consist of different components, e.g., two networks that parallelly~\cite{han2018co} or sequentially~\cite{patrini2017making} train each other via stochastic gradient descent. In the inference phase, images are then fed into the trained network, and the prediction accuracy is computed based on true labels. The core idea behind such a learning process is to filter out the negative impact of noisy labels during training and learn a model from \emph{clean} information. 

In contrast to indirectly learn the label noise dynamics, our core idea is to leverage noisy labels as part of the training and inference input, as shown in  Fig.~\ref{fig:TTmodels} (including green elements), to directly learn the noisy label dynamics and incorporate that as auxiliary input into the training process. To such an end, the ground truth of labels is assumed from human experts or oracles and provided as part of the training input. Essentially, the networks are trained by three inputs: images, their noisy labels, and the ground truth labels. 
Afterward, the trained network will be tested on images and their noisy labels. The classifier can then classify images based on image inputs and limited label info.

\section{Related Work} 

The central theme of learning on noisy labels is to increase the robustness of networks by distilling the impact of noisy labels during training.
We categorize the proposed solutions into three approaches. In addition, we review adversarial learning.

{\bf Smart sampling selection.} An intuitive solution is to select clean samples or to cleanse the (noisy) labels such that models are trained on selective clean labels~\cite{reed2014training,ren2018learning}. MentorNet~\cite{jiang2017mentornet} pre-trains a neural network as a teacher and provides a curriculum (sample weighting scheme) for the student network to select clean samples. The curriculum is dynamically learned from the data, instead of pre-defined by human experts. Decoupling~\cite{malach2017decoupling} designs an algorithm that selectively updates the weights of two base deep networks, only when there is disagreement between the two networks' predictions. Co-teaching~\cite{han2018co} and Co-teaching$+$ ~\cite{yu2019does} relies on two neural networks teaching each other the curriculum for clean data selection and training. Such approaches are inherently limited by the sample-selection bias. 

{\bf Modifying loss function.} Designing loss function that is tolerant to label noise can fundamentally strengthen the robustness of the classifier.
\cite{natarajan2013learning} derives an unbiased estimator of any loss functions for binary classification, e.g., SVM and logistic regression, in the presence of random classification noise. A simple weighted loss function is proposed to differentiate the noisy and clean label distributions, where the weights are label-dependent. Ghosh et. al~\cite{ghosh2017robust} extend the necessary condition of unbiased estimator of loss function~\cite{natarajan2013learning} from binary classification to multi-class classification. D2L~\cite{ma2018dimensionality} uses Local Intrinsic Dimensionality (LID) score as a proxy to indicate when the memorization effect on dirty labels becomes dominant. 
\cite{zhang2018generalized} proposes to incorporate use mean absolute error as a generalization to the cross entropy loss function for deep neural networks.

{\bf Noise transition model.} Incorporating the model of the noise patterns as part of the training process of the classifier is a semi-supervised approach for estimating the label quality. 
Patrini et. al~\cite{patrini2017making} estimate the noise transition matrix by manual labeling and try to minimize the distance between classification outputs and transition matrix. \cite{jindal2016learning} connects the last layer of softmax with the given labels and indirectly learns the patterns of noisy labels. Different from the other studies, \cite{hendrycks2018using} utilizes the trusted data (around 10\% of training) by introducing a loss correction technique and estimating a label corruption matrix.

{\bf Adversarial training with noisy labels}
The essence of adversarial learning is to train networks with adversarial examples, such as noisy images, so that trained networks can defend against unseen adversaries. Such learning strategies~\cite{goodfellow2014explaining} are commonly realized by a min-max game between attackers and classifiers. The adversarial images can be iteratively generated by another network~\cite{tramer2017ensemble,madry2017towards} or disturbed by Gaussian noises~\cite{kannan2018adversarial}, so called zero-knowledge attacks. Most existing adversarial learning methods focus on corrupted images, except~\cite{thekumparampil2018robustness,kaneko2019label} that focus on corrupted labels via conditional generative adversarial networks.

To the best of our knowledge, adversarial learning is yet to be explored in the problem space of noisy labels. \system presents a novel learning framework to explore the potential of adversarial learning in classification problems encountering noisy labels.



\section{Methodology}
Consider the classification problem having training set $\mathcal{D} = \{(\boldsymbol{x}_1,  \boldsymbol{y}_1, \boldsymbol{t}_1), (\boldsymbol{x}_2,  \boldsymbol{y}_2, \boldsymbol{t}_2), ... , (\boldsymbol{x}_N, \boldsymbol{y}_N, \boldsymbol{t}_N)\}$ where $\boldsymbol{x}_i$ denotes the $i^{th}$ observed sample, and $\boldsymbol{t}_i \in \{0,1\}^K$ and $\boldsymbol{y}_i \in \{0,1\}^K$ the corresponding label vectors over $K$ classes representing the clean ground truth and noisy given classes, respectively. Traditional classification problems only use the sample $\boldsymbol{x}_i$ and its true label $\boldsymbol{t}_i$. However, real-world datasets are typically affected to various degrees by label noise. Hence, for some samples, the given label $\boldsymbol{y}_i$ is different from the true label $\boldsymbol{t}_i$ even in the training set. The core contribution of the paper is a model, named \system, which leverages both $\boldsymbol{x}_i$ and $\boldsymbol{y}_i$ to predict $\boldsymbol{t}_i$, instead of only $\boldsymbol{x}_i$. Using this additional information enables the model to significantly boost the accuracy as demonstrated in \S\ref{sec:evaluation}.


\begin{figure}[t]
	\includegraphics[width=0.6\linewidth]{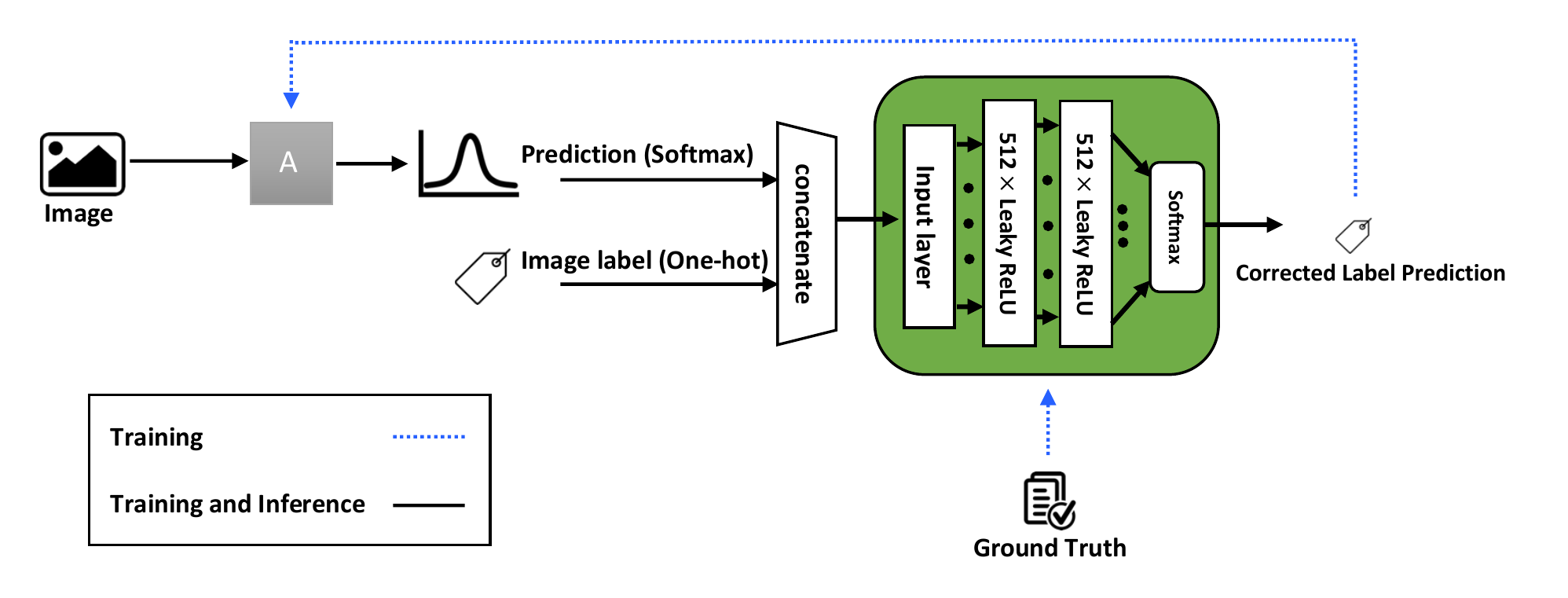}
	\centering
	\caption{\system Architecture: \ANet v.s. \ENet}
	\label{fig:network}
\end{figure}

\subsection{\system Architecture}
We address this problem via \system comprising two neural networks complete with a traditional softmax output layer named \ANet $\mathcal{A}$ and \ENet $\mathcal{E}$. Figure~\ref{fig:network} shows how the two networks are interconnected. The goal of the \ANet is to predict the label  $\boldsymbol{\hat{y}}^\mathcal{A}_i$ of an observed sample $\boldsymbol{x}_i$ while \ENet aims at correcting, if necessary, this prediction based on the output of the \ANet and the given label $\boldsymbol{y}_i$. The label corrected by the \ENet $\boldsymbol{\hat{y}}^\mathcal{E}_i$ is provided as feedback to the \ANet during training closing the loop.

\ENet acts as a supervisor which reviews and corrects the predictions of \ANet by comparing it to another label source: the given labels $\boldsymbol{y}$. We need \ENet because both label sources are affected by errors stemming from an imperfect model for the former and from label noise for the latter. From this point of view, we can consider the given labels $\boldsymbol{y}$ as the output of a second external independent imperfect model which is prone to make different errors than \ANet. The idea of having \ENet is to leverage not only the intrinsic properties of single models, as most related work does, but also the differences across the models. In the simplest case, if model A is good at classifying dogs and model B in classifying cats, we could learn to trust more model A when predicting dogs and model B when predicting cats. However, cases are rarely as easy, and we resort to the \ENet model to learn these patterns. 

To decide the type of information to exchange between \ANet and \ENet, we consider that the output layer of neural networks is traditionally a softmax transformation $\sigma(z_k)={e^{z_k}}/{\sum_{j=i}^K e^{z_j}}$. This ensures that the output vector elements are all in the range $z_j \in {0,1}, j = 1 ... K$ and their sum is $\sum_{j=1}^K z_j = 1$ satisfying the properties of a probability distribution. This probability distribution is more informative about the correctness of the prediction~\cite{liu2019zk} because it intrinsically includes information on how confident, i.e. how sharp, or how insecure, i.e. spread out, the model is on the prediction of the most likely class. Hence, we use this as the input to \ENet from \ANet rather than the sole predicted class.

The task of \ANet is to classify images. This fits well the classic state-of-the-art DNN vision-models. In our evaluation, we use the CNN defined in~\cite{wang2018iterative} having three blocks of two convolutional layers plus one pooling layer followed by a fully connected layer and the softmax output layer. Instead, the task of \ENet is to decide the correct label based on the concatenation of the class probability and given label vectors. Here we use a shallower multilayer perceptron. More details on the two submodels of \system are given in \S\ref{sec:evaluation}.

\subsection{\system Training}
Let $g^\mathcal{A}(\cdot;\boldsymbol{\phi})$ parameterized by $\boldsymbol{\phi}$ and $g^\mathcal{E}(\cdot;\boldsymbol{\omega})$ parameterized by $\boldsymbol{\omega}$ be the prediction functions. $g^\mathcal{A}()$ and $g^\mathcal{E}()$ output the class probabilities predicted by the final softmax layer of \ANet and \ENet, respectively. The training loss functions can be written as follows:
\begin{equation} \label{eq:1}
    l^\mathcal{A} = \min_{\boldsymbol{\phi}} \sum_{i=1}^{N} \mathcal{L}(g^\mathcal{E}(<g^\mathcal{A}(\boldsymbol{x}_i), \boldsymbol{y}_i>;\boldsymbol{\omega}),  g^\mathcal{A}(\boldsymbol{x}_i;\boldsymbol{\phi}))
\end{equation}
\begin{equation} \label{eq:2}
   l^\mathcal{E} = \min_{\boldsymbol{\omega}}  \sum_{i=1}^{N}
\mathcal{L}(\boldsymbol{t}_i, g^\mathcal{E}(<g^\mathcal{A}(\boldsymbol{x}_i), \boldsymbol{y}_i>;\boldsymbol{\omega})) 
\end{equation}
where $<\cdot,\cdot>$ is the concatenation function of two vectors and $\mathcal{L}$ the loss over the $K$ classes. For both networks we use $\mathcal{L}$ equal to the cross-entropy loss
fitting well the probabilistic output of softmax layer. $\mathcal{L}$ increases as predicted probability diverges from expected label.

To train the model we use the alternating minimization approach on batches of data. We first train the \ENet based on the output of the \ANet then the \ANet based on the feedback from the \ENet. Algorithm~\ref{alg:training} details this process. After random initialization of the weights $\boldsymbol{\phi}$ and $\boldsymbol{\omega}$ (Step 1) for each training step and data batch, we use $\mathcal{A}$ to predict the labels $\boldsymbol{\hat{y}}^{\mathcal{A}}$ of the observed images $\boldsymbol{x}$ (Step 4). $\boldsymbol{\hat{y}}^{\mathcal{A}}$ is concatenated with the given labels $\boldsymbol{y}$ (Step 5) as input to train $\mathcal{E}$ together with the true labels $\boldsymbol{t}$ (Step 6). After that in turn we use $\mathcal{E}$ to predict the corrected labels $\boldsymbol{\hat{y}}^{\mathcal{E}}$ (Step 7) and train $\mathcal{A}$ based on the pair $(\boldsymbol{x}, \boldsymbol{\hat{y}}^{\mathcal{E}})$ (Step 8). We use stochastic gradient descent with momentum and learning rate decay to update $\boldsymbol{\omega}$ and $\boldsymbol{\phi}$.

\begin{algorithm}[tb]
\caption{Training \system}\label{alg:training}
\SetAlgoLined
\SetKwInOut{Input}{Input}\SetKwInOut{Output}{Output}
\Input{Training set $\mathcal{D}$ made of: Observed samples $\boldsymbol{x}$, Given labels $\boldsymbol{y}$, True labels $\boldsymbol{t}$}
\Output{Trained \ANet $\mathcal{A}$ and \ENet $\mathcal{E}$}
Initialize $\mathcal{A}$ and $\mathcal{E}$ with random $\boldsymbol{\phi}$ and $\boldsymbol{\omega}$

\For{training iteration}{
  \For{each batch $B\{\boldsymbol{x},\boldsymbol{y},\boldsymbol{t}\}$ from $\mathcal{D}$}{
    $\boldsymbol{\hat{y}}^{\mathcal{A}}$ := Predict label probabilities of $\boldsymbol{x}$ by $\mathcal{A}$
    
    $\boldsymbol{z}$ := \textbf{concatenate} $<\boldsymbol{\hat{y}^{\mathcal{A}}} , \boldsymbol{y}>$
    
    Train $\mathcal{E}$ with pair $(\boldsymbol{z}, \boldsymbol{t})$ updating $\omega$
    
    $\boldsymbol{\hat{y}^{\mathcal{E}}}$ := Predict corrected label probabilities from $\boldsymbol{z}$ by $\mathcal{E}$
    
    Train $\mathcal{A}$ with pair $(\boldsymbol{x}, \boldsymbol{\hat{y}^{\mathcal{E}}})$ updating $\phi$
    
    }
  }  

\end{algorithm}


\section{Evaluation}
\label{sec:evaluation}

\subsection{Experiments Setup}
{\bf Datasets.} Our evaluation is based on three benchmarking datasets: CIFAR-10~\cite{kriz-cifar10}, CIFAR-100~\cite{kriz-cifar100} and Clothing1M~\cite{xiao2015learning}.
CIFAR-10 and CIFAR-100 include $32 \times 32$-pixel color images organized in 10 and 100 classes, respectively. The image classes range from animals to vehicles. Both datasets contain 50000 training and 10000 validation images.
Clothing1M contains images scrapped from the Internet classified into 14 classes based on the surrounding text. It is representative of real world noise (average noise rate of 39.5\%). Here we use the cleansed training, validation and testing sets of 47K, 14K and 10K samples, respectively.

{\bf Label noise.} For CIFAR-10 and CIFAR-100 we use the original labels as true labels $\boldsymbol{t}$. We generate the noisy given labels $\boldsymbol{y}$ by injecting symmetric label noise where the original label is flipped to one of the other classes with uniform probability. We use different noise ratios corresponding to flipping probabilities of $0.2$, $0.3$, $0.4$, and $0.5$. Such generating principles are applied for both training and inferences images.
Since the ground truth of 1 million training image labels in Clothing1M is not available, we use cleansed labels (47K samples) available in the dataset and then generate given (noisy) labels by using the estimated noise confusion matrix which is provided by~\cite{xiao2015learning}.

{\bf \system parameters.} For CIFAR-10 and CIFAR-100 \ANet is the 12-layer CNN architecture used in~\cite{wang2018iterative} with ReLU activation functions. \ENet is a feed-forward 4-layer neural network with Leaky ReLU activation functions in the hidden layers and sigmoid in the last layer. Both networks are implemented using Keras v2.2.4 and Tensorflow v1.13 and trained using stochastic gradient descent with momentum $0.9$, weight decay $10^{-4}$, and learning rate $0.01$. We train our model for $120$ and $200$ epochs for CIFAR-10 and CIFAR-100, respectively. For Clothing1M, we resize each image to $256 \times 256$ pixels and crop the center to $224 \times 224$.  We use ResNet50 for \ANet with SGD optimizer and momentum of 0.9. The weight decay factor is $5 \times 10^{-3}$, and the batch size is 16. The initial learning rate is 0.002 and decreased by 10 every 5 epochs. The total training epochs are 50. The \ENet architecture remains the same. All experiments run on servers equipped with 8-cores @ 2.4GHz, 64GB of RAM, and an NVIDIA TITAN X GPU.

\subsection{Competing Methods} We consider the following four methods, which aim to filter out the (impact of) noisy labels by altering the loss function, selecting the clean labels, and inferring the noise transition matrix. Competing models are based on their original code and settings.  

\setlist{nolistsep}
\begin{itemize}[noitemsep]
\item[]{\bf D2L}~\cite{ma2018dimensionality}: uses the Local Intrinsic Dimensionality (LID) to detect points of noisy data and modifies the loss function based on the LID score.
\item[]{\bf Co-teaching}~\cite{han2018co}: uses two neural networks to teach each other by selecting and exchanging the more informative data batches where the selection leverages the memory effect of neural networks.
\item[]{\bf Bootstrap}~\cite{reed2014training}: uses a weighted combination of the original label and prediction of the model as the final prediction. 
\item[]{\bf Forward}~\cite{patrini2017making}: uses the noise transition matrix to correct the labels before training.
\end{itemize}

{\bf Comparison modes}
For a fair comparison, we compare the competing models against \system under two scenarios: without and with using the given labels $\boldsymbol{y}$ during inference. In the case without $\boldsymbol{y}$, the predictions are taken from \ANet, and \ENet is used only during training. In the case with $\boldsymbol{y}$, we consider the whole \system, including \ENet in both training and inference, and the predictions are taken from \ENet. Additionally, we evaluate the effect of decreasing amounts of training data which range from randomly selected $100\%$ to $20\%$ of the training samples for each dataset. Experiment across competing models all use the same training and validation sets.

{\bf Metrics of interests}
We present the inference accuracy of \system and all four competing methods.
As a performance metric, we use the accuracy evaluated on the validation data computed as the ratio of the number of correct predictions, i.e. equal to the original true labels $\boldsymbol{t}$, divided by the total number of validation samples.
For \system, we present two sets of accuracy results from \system: one is based on the inference from the sole trained \ANet network, and the other is based on the joint inference from \ANet and \ENet. Such convention are used in Table~\ref{tab:cifar10n} and~\ref{tab:cifar100n}. The difference between the accuracy values represents the auxiliary learning capacity of \ENet.

\subsection{Results}

{\bf CIFAR-10}
We report accuracy results under all noise levels and amounts of training data in Table~\ref{tab:cifar10n}. Starting with 100\% training data, one can see that \system achieves the accuracy of 89.23\%, 88.30\%, 84.36\%, and 80.73\% for the cases of 20\%, 30\%, 40\%, and 50\% noise ratio. These results are significantly better than the competing image-only models. The accuracy of \system is 1.79\% to 11.92\% higher under all considered noise ratios.
We attribute the superior accuracy to \ENet that leverages well the information from both \ANet and the given labels. If the given labels are not available, the simpler \ANet still benefits from the feedback of \ENet during training. As a result, even the simpler \ANet is able to compete well and surpass some of the competing models. For the same scenarios, \ANet alone still achieves the accuracy of 82.29\%, 81.85\%, 79.53\%,  and 76.45\% consistently beating Forward and Bootstrap and sometimes Co-teaching.

In case of reduced training data, the accuracy of \system decreases with the amount of data. However, \system is still the best across all competing models.
Even with more training data, e.g. 60\% data and 20\% noise, we achieve 86.27\% and 79.18\% accuracy in \ENet and \ANet, respectively.
The best rival achieves only 79.73\%. In an extreme case, with only 20\% training data, \ENet is significantly better than all others.
For fair comparison, we summarize the training data required by \system to reach similar or higher accuracy than the four competing methods each trained with 100\% training data. \system achieves this using only 40\%, 40\%, 60\%, and 80\% training data under 20\%, 30\%, and 40\% and 50\% noise, respectively.
One effect of diminishing training data is that the difference between \system and \ANet becomes more significant. We interpret this as the lower the amount of available data, the more one should use any possibly source of information, e.g., the given labels, even during testing.
The training loss for both \ANet and \ENet converge to a lower bound, but the evolution of \ENet loss is smoother than its \ANet equivalent. This indicates that \system makes it easier for \ENet to correct labels than for \ANet to classify images. 

\begin{table}
\centering
\caption {Inference accuracy on CIFAR-10} \label{tab:cifar10n} 
\resizebox{1.0\linewidth}{!}{%
\begin{tabular}{|@{}c@{}|@{}c@{}|@{}c@{}|c|c|@{}c@{}|@{}c@{}|@{}c@{}|@{}c@{}|c|c|@{}c@{}|@{}c@{}|@{}c@{}|} 
\cline{3-14}
\multicolumn{2}{c|}{}  & \multicolumn{6}{|c|}{Noise Ratio = 20\%}                                                                                                                                      & \multicolumn{6}{c|}{Noise Ratio = 40\%}                                                                                                         \\ 
\cline{3-14}
\multicolumn{2}{c|}{} & \multicolumn{2}{c|}{ExpertNet} & \multirow{2}{*}{D2L} & \multirow{2}{*}{Co-Teach.} & \multirow{2}{*}{Bootstrap} & \multirow{2}{*}{Forward} & \multicolumn{2}{c|}{ExpertNet} & \multirow{2}{*}{D2L} & \multirow{2}{*}{Co-Teach.} & \multirow{2}{*}{Bootstrap} & \multirow{2}{*}{Forward}  \\ 
\cline{3-4}\cline{9-10}
\multicolumn{2}{c|}{}                            & Amateur~ & Expert              &                      &                              &                            &                          & Amateur~ & Expert              &                      &                               &                             &                           \\ 
\hline
\multirow{5}{*}{\rotatebox[origin=c]{90}{Training data}} & 100\%                          & 82.29    & \textbf{89.23}      & 84.75                & 82.45                        & 81.80                      & 83.11                    & 79.53~   & \textbf{84.36}      & 80.69                & 77.28                         & 72.44                       & 78.12                     \\
 & 80\%                           & 81.74    & \textbf{88.61}      & 82.85                & 81.57                        & 79.98                      & 81.43                    & 76.16    & \textbf{81.76}      & 79.11                & 75.74                         & 71.50                       & 72.76                     \\
 & 60\%                           & 79.18    & \textbf{86.27}      & 79.73                & 79.30                        & 74.19                      & 75.40                    & 73.35    & \textbf{79.33}      & 76.63                & 73.95                         & 58.02                       & 55.67                     \\
 & 40\%                           & 75.11    & \textbf{86.01}      & 77.94                & 77.09                        & 63.82                      & 60.31                    & 67.32    & \textbf{74.78}      & 71.84                & 73.95                         & 25.11                       & 31.88                     \\
 & 20\%                           & 67.45    & \textbf{82.23}      & 70.47                & 70.37                        & 23.82                      & 31.35                    & 56.33    & 67.88               & \textbf{67.96}       & 63.08                         & 18.91                       & 21.16                     \\ 
\hline
\multicolumn{2}{c|}{} & \multicolumn{6}{|c|}{Noise Ratio = 30\%}                                                                                                                                      & \multicolumn{6}{c|}{Noise Ratio = 50\%}                                                                                                         \\ 
\hline
\multirow{5}{*}{\rotatebox[origin=c]{90}{Training data}} & 100\%                          & 81.85    & \textbf{88.30}      & 82.45                & 80.29                        & 77.14                      & 81.68                    & 76.45    & \textbf{80.73}      & 78.94                & 74.47                         & 70.14                       & 76.23                     \\
 & 80\%                           & 79.87    & \textbf{85.83}      & 81.27                & 79.16                        & 75.60                      & 79.38                    & 72.24    & \textbf{78.48}      & 76.43                & 71.52                         & 57.84                       & 63.33                     \\
 & 60\%                           & 77.73    & \textbf{86.32}      & 79.14                & 76.87                        & 70.09                      & 68.35                    & 69.91    & \textbf{75.06}      & 73.73                & 68.80                         & 31.98                       & 37.01                     \\
 & 40\%                           & 71.88    & \textbf{82.12}      & 75.68                & 72.87                        & 35.13                      & 37.87                    & 61.86    & \textbf{69.34}      & 68.63                & 64.25                         & 22.95                       & 26.87                     \\
 & 20\%                           & 61.06    & \textbf{75.44}      & 70.78                & 67.37                        & 20.99                      & 28.99                    & 50.96    & \textbf{61.37}      & 59.08                & 57.11                         & 15.63                       & 17.89                     \\
\hline
\end{tabular}
}
\end{table}






{\bf CIFAR-100} is significantly more difficult than CIFAR-10. First, the number of classes increase by a factor 10. Second, the training samples per class reduce by a factor 10. Consequently, the achieved accuracy scores shown in Table~\ref{tab:cifar100n} are lower. Even so \system achieves 86.72\%, 79.92\%, 73.87\%, and 66.11\% for noise ratios of 20\%, 30\%, 40\%, and 50\%, respectively. Moreover, the advantage of using \ENet as guidance is more pronounced. Not only no other model except \system is able to reach 60\% accuracy, but also \ANet consistently reaches similar performance as D2L except under amounts of training data below 60\%. However, \ANet is in line with Forward. Another positive result is that \system seems to be the least affected by diminishing training data. Unfortunately, the same does not hold for increasing noise levels. 
For a fair comparison, we identify the minimum required training data for \system to achieve similar or higher accuracy as the other four approaches that leverage 100\% training data. Under all noise ratios, \system only needs to have 20\% training data to outperform other methods by large margins, ranging from 15\% to 27\% absolute accuracy improvement.
\ENet again has a smoother loss evolution than \ANet, but the loss evolution of \ANet on CIFAR-100 is spikier than on CIFAR-10, implying the increased difficulty of \ANet to differentiate 100 classes.


\begin{table}
\centering
\caption {Inference accuracy on CIFAR-100} \label{tab:cifar100n} 
\resizebox{1.0\linewidth}{!}{%
\begin{tabular}{|@{}c@{}|@{}c@{}|@{}c@{}|c|c|@{}c@{}|@{}c@{}|@{}c@{}|@{}c@{}|c|c|@{}c@{}|@{}c@{}|@{}c@{}|} 
\cline{3-14}
\multicolumn{2}{c|}{} & \multicolumn{6}{|c|}{Noise Ratio = 20\%}                                                                                                                                      & \multicolumn{6}{c|}{Noise Ratio = 40\%}                                                                                                         \\ 
\cline{3-14}
\multicolumn{2}{c|}{} & \multicolumn{2}{c|}{ExpertNet} & \multirow{2}{*}{D2L} & \multirow{2}{*}{Co-Teach.} & \multirow{2}{*}{Bootstrap} & \multirow{2}{*}{Forward} & \multicolumn{2}{c|}{ExpertNet} & \multirow{2}{*}{D2L} & \multirow{2}{*}{Co-Teach.} & \multirow{2}{*}{Bootstrap} & \multirow{2}{*}{Forward}  \\ 
\cline{3-4}\cline{9-10}
\multicolumn{2}{c|}{}                                & Amateur & Expert               &                      &                              &                            &                          & Amateur~ & Expert              &                      &                               &                             &                           \\ 
\hline
\multirow{5}{*}{\rotatebox[origin=c]{90}{Training data}} & 100\%                          & 59.24   & \textbf{86.72}       & 55.70                & 52.74                        & 52.58                      & 59.87                    & 53.04    & \textbf{73.87}      & 49.50                & 41.87                         & 40.11                       & 49.44                     \\
 & 80\%                           & 54.56   & \textbf{85.38}~      & 51.26                & 50.01                        & 48.95                      & 55.51                    & 49.91    & \textbf{71.06}      & 45.17                & 39.88                         & 37.24                       & 47.23                     \\
 & 60\%                           & 50.01   & \textbf{84.85}       & 48.33                & 42.82                        & 41.62                      & 50.76                    & 41.76    & \textbf{68.80}      & 39.89                & 33.54                         & 30.63                       & 44.22                     \\
 & 40\%                           & 44.13   & \textbf{82.51}       & 42.48                & 36.75                        & 32.68                      & 48.04                    & 34.96    & \textbf{64.33}      & 36.82                & 28.33                         & 22.77                       & 41.35                     \\
 & 20\%                           & 31.11   & \textbf{80.74}       & 31.19                & 27.93                        & 24.01                      & 32.65                    & 23.12    & \textbf{61.89}      & 24.33                & 19.92                         & 18.45                       & 26.66                     \\ 
\hline
\multicolumn{2}{c|}{} & \multicolumn{6}{|c|}{Noise Ratio = 30\%}                                                                                                                                      & \multicolumn{6}{c|}{Noise Ratio = 50\%}                                                                                                         \\ 
\hline
\multirow{5}{*}{\rotatebox[origin=c]{90}{Training data}} & 100\%                          & 56.78   & \textbf{79.92}       & 51.13                & 45.68                        & 44.99                      & 54.18                    & 48.65    & \textbf{66.11}      & 43.56                & 35.89                         & 39.84                       & 46.06                     \\
 & 80\%                           & 52.98   & \textbf{78.61}       & 48.26                & 44.36                        & 41.47                      & 53.66                    & 43.61    & \textbf{63.45}      & 37.98                & 33.69                         & 33.79                       & 41.23                     \\
 & 60\%                           & 46.34   & \textbf{76.18}       & 43.79                & 39.71                        & 35.46                      & 48.46                    & 35.55    & \textbf{58.56}      & 33.37                & 29.14                         & 28.98                       & 37.05                     \\
 & 40\%                           & 39.87   & \textbf{73.05}       & 40.28                & 33.83                        & 30.21                      & 47.71                    & 25.51    & \textbf{53.02}      & 30.76                & 23.96                         & 15.80                       & 34.27                     \\
 & 20\%                           & 26.45   & \textbf{71.72}       & 27.98                & 24.21                        & 21.78                      & 29.77                    & 16.87    & \textbf{51.01}      & 17.54                & 16.85                         & 10.54                       & 21.87                     \\
\hline
\end{tabular}
}
\end{table}

\system achieves remarkable inference accuracy in the presence of noise labels on significantly smaller sets of training data, compared to state-of-the-art methods. The effective design of \system is particularly evident for more difficult benchmarks, such as CIFAR-100. The combined architecture of \ANet and \ENet outperforms other methods even when learning from just 20\% of training data used for the others.


\begin{table}
\centering
\caption{Inference accuracy on Clothing1M (affected by real world noise)} \label{tab:clothing1M}
\resizebox{0.8\linewidth}{!}{%
\begin{tabular}{|l|c|c|c|c|c|} 
\cline{2-6}
\multicolumn{1}{c|}{} & ExpertNet      & D2L   & Co-Teaching & Forward & Bootstrap  \\ 
\hline
Training data = 50\%                                 & \textbf{69.83} & 49.05 & 50.11       & 51.26   & 48.94      \\ 
\hline
Training data = 100\%                                & \textbf{83.42} & 69.43 & 69.92       & 70.04   & 68.77      \\
\hline
\end{tabular}
}
\end{table}


{\bf Clothing1M}
We summarize results in Table~\ref{tab:clothing1M} with full training data and randomly selected 50\% data. When using all the training set, \system achieves 83.42\% accuracy, which is 13 points higher than the second best approach, i.e., forward at 70.04\%. This is due to the capacity of \ENet to learn the real world noise pattern. When halving the training set, \system still achieves 69.83\%, which is roughly the result the four competing methods reach using 100\% training data. In other words, having the ground truth for half of the data \system can outperform other approaches which do not leverage the knowledge of noise patterns. \system has the best relative performance on CIFAR-100, followed by Clothing1M, and CIFAR-10, reflecting the decreasing importance and difficulty to learn the noise patterns.
Clothing1M results further accentuate the idea of \system that learning from both images and (noisy) labels can strengthen the robustness and efficiency of deep neural networks.

\section{Conclusion}
 Motivated by the observation that images in the public domain are often bundled with pre-existing noisy labels, this paper presents a novel and effective learning paradigm, called \system, which infers the images by both inputs of images and noisy labels via two networks, i.e., \ANet and \ENet. The core idea of \system is to train \ANet, and \ENet with each other, where \ANet is a deep CNN and \ENet imitates how human experts correct the output of \ANet by the ground truth. As such, \system can effectively classify images via auxiliary information of noisy labels - proactively turning dirty labels to a learning advantage. Our empirical results show that \system can be generalized on extensive and challenging scenarios, i.e., the combinations of noise ratios and training data reduction, and significantly outperforms existing robust network classifiers on both CIFAR benchmarks and real world dataset. 



\bibliography{egbib}

\begin{thebibliography}{31}
\providecommand{\natexlab}[1]{#1}
\providecommand{\url}[1]{\texttt{#1}}
\expandafter\ifx\csname urlstyle\endcsname\relax
  \providecommand{\doi}[1]{doi: #1}\else
  \providecommand{\doi}{doi: \begingroup \urlstyle{rm}\Url}\fi

\bibitem[Blum et~al.(2003)Blum, Kalai, and Wasserman]{blum2003noise}
Avrim Blum, Adam Kalai, and Hal Wasserman.
\newblock Noise-tolerant learning, the parity problem, and the statistical
  query model.
\newblock \emph{Journal of the ACM (JACM)}, 50\penalty0 (4):\penalty0 506--519,
  2003.

\bibitem[Frome et~al.(2013)Frome, Corrado, Shlens, Bengio, Dean, Ranzato, and
  Mikolov]{Frome:NIPS13:DeVise}
Andrea Frome, Gregory~S. Corrado, Jonathon Shlens, Samy Bengio, Jeffrey Dean,
  Marc'Aurelio Ranzato, and Tomas Mikolov.
\newblock Devise: {A} deep visual-semantic embedding model.
\newblock In \emph{Advances in Neural Information Processing System}, pages
  2121--2129, 2013.

\bibitem[Ghosh et~al.(2017)Ghosh, Kumar, and Sastry]{ghosh2017robust}
Aritra Ghosh, Himanshu Kumar, and PS~Sastry.
\newblock Robust loss functions under label noise for deep neural networks.
\newblock In \emph{Thirty-First AAAI Conference on Artificial Intelligence},
  pages 1919--1925, 2017.

\bibitem[Goodfellow et~al.(2015)Goodfellow, Shlens, and
  Szegedy]{goodfellow2014explaining}
Ian Goodfellow, Jonathon Shlens, and Christian Szegedy.
\newblock Explaining and harnessing adversarial examples.
\newblock In \emph{International Conference on Learning Representations}, 2015.
\newblock URL \url{http://arxiv.org/abs/1412.6572}.

\bibitem[Han et~al.(2018)Han, Yao, Yu, Niu, Xu, Hu, Tsang, and
  Sugiyama]{han2018co}
Bo~Han, Quanming Yao, Xingrui Yu, Gang Niu, Miao Xu, Weihua Hu, Ivor Tsang, and
  Masashi Sugiyama.
\newblock Co-teaching: Robust training of deep neural networks with extremely
  noisy labels.
\newblock In \emph{Advances in Neural Information Processing Systems}, pages
  8527--8537, 2018.

\bibitem[Hendrycks et~al.(2018)Hendrycks, Mazeika, Wilson, and
  Gimpel]{hendrycks2018using}
Dan Hendrycks, Mantas Mazeika, Duncan Wilson, and Kevin Gimpel.
\newblock Using trusted data to train deep networks on labels corrupted by
  severe noise.
\newblock In \emph{Advances in neural information processing systems}, pages
  10456--10465, 2018.

\bibitem[Jiang et~al.(2018)Jiang, Zhou, Leung, Li, and
  Fei{-}Fei]{jiang2017mentornet}
Lu~Jiang, Zhengyuan Zhou, Thomas Leung, Li{-}Jia Li, and Li~Fei{-}Fei.
\newblock Mentornet: Learning data-driven curriculum for very deep neural
  networks on corrupted labels.
\newblock In \emph{International Conference on Machine Learning (ICML)}, pages
  2309--2318, 2018.

\bibitem[Jindal et~al.(2016)Jindal, Nokleby, and Chen]{jindal2016learning}
Ishan Jindal, Matthew Nokleby, and Xuewen Chen.
\newblock Learning deep networks from noisy labels with dropout regularization.
\newblock In \emph{IEEE International Conference on Data Mining (ICDM)}, pages
  967--972, 2016.

\bibitem[Kaneko et~al.(2019)Kaneko, Ushiku, and Harada]{kaneko2019label}
Takuhiro Kaneko, Yoshitaka Ushiku, and Tatsuya Harada.
\newblock Label-noise robust generative adversarial networks.
\newblock In \emph{IEEE Conference on Computer Vision and Pattern Recognition
  (CVPR)}, pages 2467--2476, 2019.

\bibitem[Kannan et~al.(2018)Kannan, Kurakin, and
  Goodfellow]{kannan2018adversarial}
Harini Kannan, Alexey Kurakin, and Ian Goodfellow.
\newblock Adversarial logit pairing.
\newblock \emph{arXiv preprint arXiv:1803.06373}, 2018.

\bibitem[Krizhevsky et~al.({\natexlab{a}})Krizhevsky, Nair, and
  Hinton]{kriz-cifar10}
Alex Krizhevsky, Vinod Nair, and Geoffrey Hinton.
\newblock Cifar-10 (canadian institute for advanced research).
\newblock {\natexlab{a}}.
\newblock URL \url{http://www.cs.toronto.edu/~kriz/cifar.html}.

\bibitem[Krizhevsky et~al.({\natexlab{b}})Krizhevsky, Nair, and
  Hinton]{kriz-cifar100}
Alex Krizhevsky, Vinod Nair, and Geoffrey Hinton.
\newblock Cifar-100 (canadian institute for advanced research).
\newblock {\natexlab{b}}.
\newblock URL \url{http://www.cs.toronto.edu/~kriz/cifar.html}.

\bibitem[Kurakin et~al.(2016)Kurakin, Goodfellow, and
  Bengio]{kurakin2016adversarial}
Alexey Kurakin, Ian Goodfellow, and Samy Bengio.
\newblock Adversarial machine learning at scale.
\newblock \emph{arXiv preprint arXiv:1611.01236}, 2016.

\bibitem[Liu et~al.(2019)Liu, Khalil, and Khreishah]{liu2019zk}
Guanxiong Liu, Issa Khalil, and Abdallah Khreishah.
\newblock Zk-gandef: {A} {GAN} based zero knowledge adversarial training
  defense for neural networks.
\newblock In \emph{{IEEE/IFIP} International Conference on Dependable Systems
  and Networks (DSN)}, pages 64--75, 2019.

\bibitem[Madry et~al.(2018)Madry, Makelov, Schmidt, Tsipras, and
  Vladu]{madry2017towards}
Aleksander Madry, Aleksandar Makelov, Ludwig Schmidt, Dimitris Tsipras, and
  Adrian Vladu.
\newblock Towards deep learning models resistant to adversarial attacks.
\newblock In \emph{6th International Conference on Learning Representations
  (ICLR)}, 2018.

\bibitem[Malach and Shalev-Shwartz(2017)]{malach2017decoupling}
Eran Malach and Shai Shalev-Shwartz.
\newblock Decoupling" when to update" from" how to update".
\newblock In \emph{Advances in Neural Information Processing Systems}, pages
  960--970, 2017.

\bibitem[Mirza and Osindero(2014)]{mirza2014conditional}
Mehdi Mirza and Simon Osindero.
\newblock Conditional generative adversarial nets.
\newblock \emph{arXiv preprint arXiv:1411.1784}, 2014.

\bibitem[Natarajan et~al.(2013)Natarajan, Dhillon, Ravikumar, and
  Tewari]{natarajan2013learning}
Nagarajan Natarajan, Inderjit~S Dhillon, Pradeep~K Ravikumar, and Ambuj Tewari.
\newblock Learning with noisy labels.
\newblock In \emph{Advances in neural information processing systems}, pages
  1196--1204, 2013.

\bibitem[Patrini et~al.(2017)Patrini, Rozza, Krishna~Menon, Nock, and
  Qu]{patrini2017making}
Giorgio Patrini, Alessandro Rozza, Aditya Krishna~Menon, Richard Nock, and
  Lizhen Qu.
\newblock Making deep neural networks robust to label noise: A loss correction
  approach.
\newblock In \emph{IEEE Conference on Computer Vision and Pattern Recognition
  (CVPR)}, pages 1944--1952, 2017.

\bibitem[Reed et~al.(2015)Reed, Lee, Anguelov, Szegedy, Erhan, and
  Rabinovich]{reed2014training}
Scott~E. Reed, Honglak Lee, Dragomir Anguelov, Christian Szegedy, Dumitru
  Erhan, and Andrew Rabinovich.
\newblock Training deep neural networks on noisy labels with bootstrapping.
\newblock In \emph{3rd International Conference on Learning Representations,
  (CLR) Workshop}, 2015.

\bibitem[Ren et~al.(2018)Ren, Zeng, Yang, and Urtasun]{ren2018learning}
Mengye Ren, Wenyuan Zeng, Bin Yang, and Raquel Urtasun.
\newblock Learning to reweight examples for robust deep learning.
\newblock In \emph{International Conference on Machine Learning (ICML)}, pages
  4331--4340, 2018.

\bibitem[Szegedy et~al.(2014)Szegedy, Zaremba, Sutskever, Bruna, Erhan,
  Goodfellow, and Fergus]{Szegedy2014intriguing}
Christian Szegedy, Wojciech Zaremba, Ilya Sutskever, Joan Bruna, Dumitru Erhan,
  Ian Goodfellow, and Rob Fergus.
\newblock Intriguing properties of neural networks.
\newblock In \emph{International Conference on Learning Representations
  (ICLR)}, 2014.
\newblock URL \url{http://arxiv.org/abs/1312.6199}.

\bibitem[Thekumparampil et~al.(2018)Thekumparampil, Khetan, Lin, and
  Oh]{thekumparampil2018robustness}
Kiran~K Thekumparampil, Ashish Khetan, Zinan Lin, and Sewoong Oh.
\newblock Robustness of conditional gans to noisy labels.
\newblock In \emph{Advances in Neural Information Processing Systems}, pages
  10271--10282, 2018.

\bibitem[Tram{\`e}r et~al.(2017)Tram{\`e}r, Kurakin, Papernot, Goodfellow,
  Boneh, and McDaniel]{tramer2017ensemble}
Florian Tram{\`e}r, Alexey Kurakin, Nicolas Papernot, Ian Goodfellow, Dan
  Boneh, and Patrick McDaniel.
\newblock Ensemble adversarial training: Attacks and defenses.
\newblock \emph{arXiv preprint arXiv:1705.07204}, 2017.

\bibitem[Wang et~al.(2018{\natexlab{a}})Wang, Liu, Ma, Bailey, Zha, Song, and
  Xia]{wang2018iterative}
Yisen Wang, Weiyang Liu, Xingjun Ma, James Bailey, Hongyuan Zha, Le~Song, and
  Shu-Tao Xia.
\newblock Iterative learning with open-set noisy labels.
\newblock In \emph{IEEE Conference on Computer Vision and Pattern Recognition
  (CVPR)}, pages 8688--8696, 2018{\natexlab{a}}.

\bibitem[Wang et~al.(2018{\natexlab{b}})Wang, Ma, Houle, Xia, and
  Bailey]{ma2018dimensionality}
Yisen Wang, Xingjun Ma, Michael~E Houle, Shu-Tao Xia, and James Bailey.
\newblock Dimensionality-driven learning with noisy labels.
\newblock \emph{International Conference on Machine Learning (ICML)}, pages
  3361--3370, 2018{\natexlab{b}}.

\bibitem[Xiao et~al.(2015)Xiao, Xia, Yang, Huang, and Wang]{xiao2015learning}
Tong Xiao, Tian Xia, Yi~Yang, Chang Huang, and Xiaogang Wang.
\newblock Learning from massive noisy labeled data for image classification.
\newblock In \emph{IEEE conference on computer vision and pattern recognition
  (CVPR)}, pages 2691--2699, 2015.

\bibitem[Yan et~al.(2014)Yan, Rosales, Fung, Subramanian, and
  Dy]{yan2014learning}
Yan Yan, R{\'o}mer Rosales, Glenn Fung, Ramanathan Subramanian, and Jennifer
  Dy.
\newblock Learning from multiple annotators with varying expertise.
\newblock \emph{Machine learning}, 95\penalty0 (3):\penalty0 291--327, 2014.

\bibitem[Yu et~al.(2019)Yu, Han, Yao, Niu, Tsang, and Sugiyama]{yu2019does}
Xingrui Yu, Bo~Han, Jiangchao Yao, Gang Niu, Ivor~W Tsang, and Masashi
  Sugiyama.
\newblock How does disagreement help generalization against label corruption?
\newblock In \emph{International Conference on Machine Learning (ICML)}, pages
  7164--7173, 2019.

\bibitem[Zhang et~al.(2017)Zhang, Bengio, Hardt, Recht, and
  Vinyals]{Zhang2017memorization}
Chiyuan Zhang, Samy Bengio, Moritz Hardt, Benjamin Recht, and Oriol Vinyals.
\newblock Understanding deep learning requires rethinking generalization.
\newblock In \emph{5th International Conference on Learning Representations
  (ICLR)}, 2017.

\bibitem[Zhang and Sabuncu(2018)]{zhang2018generalized}
Zhilu Zhang and Mert Sabuncu.
\newblock Generalized cross entropy loss for training deep neural networks with
  noisy labels.
\newblock In \emph{Advances in Neural Information Processing Systems}, pages
  8778--8788, 2018.

\end{thebibliography}
\end{document}